\documentclass[sigconf,natbib=true]{acmart}
\usepackage{subfigure}
\usepackage{amsmath}
\usepackage{multirow}
\usepackage{graphicx}
\usepackage{hyperref}
\usepackage{dblfloatfix}

\copyrightyear{2023} 
\acmYear{2023} 
\setcopyright{acmlicensed}\acmConference[COMPASS '23]{ACM SIGCAS/SIGCHI Conference on Computing and Sustainable Societies}{August 16--19, 2023}{Cape Town, South Africa}
\acmBooktitle{ACM SIGCAS/SIGCHI Conference on Computing and Sustainable Societies (COMPASS '23), August 16--19, 2023, Cape Town, South Africa}
\acmPrice{15.00}
\acmDOI{10.1145/3588001.3609359}
\acmISBN{979-8-4007-0149-8/23/08}

\begin{document}
\title{Poverty rate prediction using multi-modal survey and earth observation data}

\author{Simone Fobi}
\affiliation{%
  \institution{AI for Good Lab, Microsoft}
  \city{Redmond}
    \country{USA}
}
\email{sfobinsutezo@microsoft.com}
\author{Manuel Cardona}
\affiliation{%
  \institution{Innovations for Poverty Action}
  \city{Mexico City}
    \country{Mexico}
}
\email{mcardona@poverty-action.org}
\author{Elliott Collins}
\affiliation{%
  \institution{Innovations for Poverty Action}
  \city{Washington DC}
    \country{USA}
}
\email{ecollins@poverty-action.org}
\author{Caleb Robinson}
\affiliation{%
  \institution{AI for Good Lab, Microsoft}
  \city{Redmond}
    \country{USA}
}
\email{caleb.robinson@microsoft.com}

\author{Anthony Ortiz}
\affiliation{%
  \institution{AI for Good Lab, Microsoft}
  \city{Redmond}
    \country{USA}
}
\email{anthony.ortiz@microsoft.com}
\author{Tina Sederholm}
\affiliation{%
  \institution{AI for Good Lab, Microsoft}
  \city{Redmond}
    \country{USA}
}
\email{tinasepubs@outlook.com}
\author{Rahul Dodhia}
\affiliation{%
  \institution{AI for Good Lab, Microsoft}
  \city{Redmond}
    \country{USA}
}
\email{rahul.dodhia@microsoft.com}
\author{Juan Lavista Ferres}
\affiliation{%
  \institution{AI for Good Lab, Microsoft}
    \city{Redmond}
    \country{USA}
}
\email{jlavista@microsoft.com}

\renewcommand{\shortauthors}{}

\begin{abstract}
This work presents an approach for combining household demographic and living standards survey questions with features derived from satellite imagery to predict the poverty rate of a region. Our approach utilizes visual features obtained from a single-step featurization method applied to freely available 10m/px Sentinel-2 surface reflectance satellite imagery. These visual features are combined with ten survey questions in a proxy means test (PMT) to estimate whether a household is below the poverty line. We show that the inclusion of visual features reduces the mean error in poverty rate estimates from 4.09\% to 3.88\% over a nationally representative out-of-sample test set. 
In addition to including satellite imagery features in proxy means tests, we propose an approach for selecting a subset of survey questions that are complementary to the visual features extracted from satellite imagery. Specifically, we design a survey variable selection approach guided by the full survey and image features and use the approach to determine the most relevant set of small survey questions to include in a PMT. We validate the choice of small survey questions in a downstream task of predicting the poverty rate using the small set of questions. This approach results in the best performance -- errors in poverty rate decrease from 4.09\% to 3.71\%. We show that extracted visual features encode geographic and urbanization differences between regions.
\end{abstract}



\pagestyle{empty} 
\maketitle

\section{Introduction} \label{intro}
Monitoring poverty is a crucial component for achieving \textit{Sustainable Development Goal 1: No Poverty} by 2030. Global poverty rates had decreased to a low of 8.6\% in 2019, however estimates suggests that the COVID-19 pandemic contributed to a 1\% increase by the end of 2020, rewinding progress in poverty reduction~\cite{sdg1}. Thus, accurate measurements of poverty rates are needed to track progress towards poverty elimination goals and to measure the impact of poverty reduction strategies and interventions.

The \textit{poverty rate} is defined at a regional level as the proportion of people that fall below the \textit{poverty line}~\cite{povertyrate} -- a threshold that represents the minimum amount of money required to cover a household's basic needs. 
If a households' income or consumption expenditures are below this threshold, then the household is considered to be poor. Household consumption expenditure data are needed to estimate poverty rates and are calculated from representative and extensive household demographic and living standard surveys such as the Demographic and Health Surveys (DHS) and the Living Standards Measurement Study (LSMS).

While conducting extensive surveys represents an optimal approach for estimating household consumption expenditure and, by extension, the poverty rate, a single round of such surveys over an entire country is laborious and entails substantial costs~\cite{hall2022review}. The significant obstacles of high survey costs pose a major challenge, particularly for lower-income economies, in accurately assessing poverty rates over different locations and periods. Consequently, this hinders the effective implementation of poverty alleviation programs. Two sets of methodological improvements have been made to estimate poverty in an attempt to mitigate the high costs of extensive surveys: i) Proxy Means Tests (PMT) and ii) satellite imagery and machine learning based methods. PMTs utilize a small survey (10 to 30 questions) to obtain a set of measurable household variables which are in turn used for poverty estimation~\cite{kshirsagar2017household}. Here, the regression coefficients for the PMT are calibrated to the extensive survey data, and the calibrated PMT is used to make out-of-sample predictions given a much smaller survey~\cite{brown2018poor}.

More recently, machine learning methods have been used to estimate asset wealth and poverty across multiple countries using non-survey datasets such as satellite imagery ~\cite{jean2016combining,yeh2020using,head2017can,tingzon2019mapping,tang2022predicting,ni2020investigation,jarry2021assessment}, call data records~\cite{pokhriyal2017combining,steele2017mapping}, social media activity~\cite{ledesma2020interpretable} or a combination of these~\cite{chi2022microestimates}. Notably, these works show that greater than 50\% of the variation in survey-measured poverty can be explained with satellite image-based poverty predictions~\cite{hall2022review}. 

Both PMT and satellite-image based methods have strengths and weaknesses. PMTs utilize a reduced set of survey questions, thereby decreasing survey cost and time, but necessarily discard additional relevant information for improved poverty estimates. Satellite-image based methods utilize readily available data with large geographic and temporal coverage to estimate poverty but can only incorporate information that is readily available in the imagery.

\begin{figure*}[th]
\centering
\includegraphics[width=0.9\textwidth]{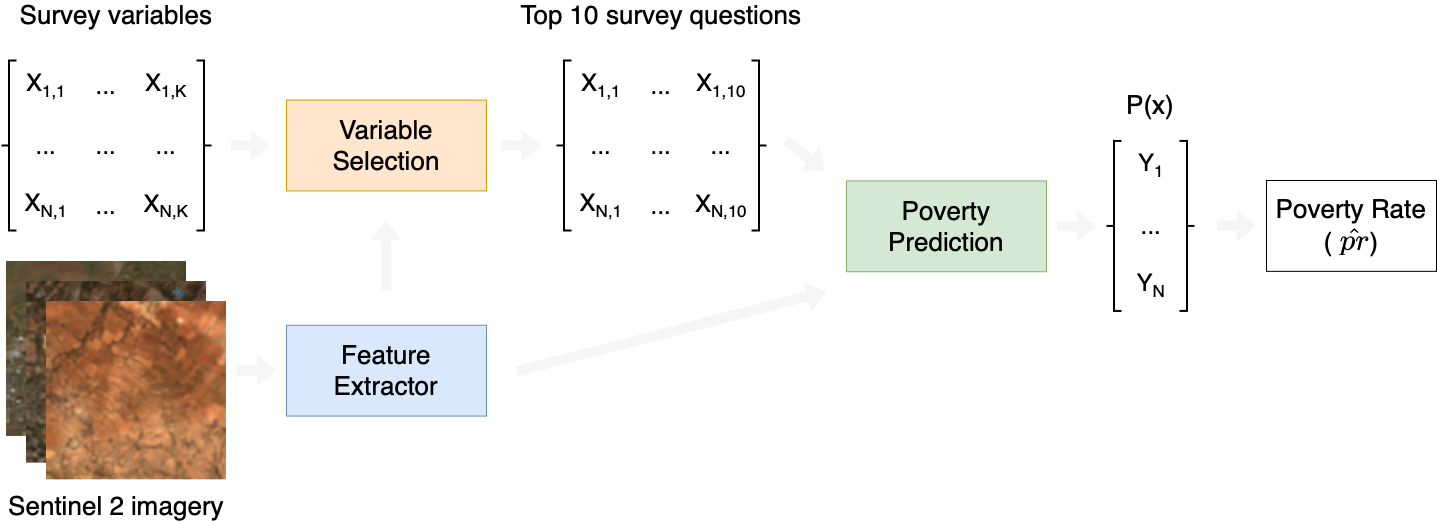}
\caption{Illustration of our methodology that combines satellite imagery in a proxy means test to estimate poverty rate. \textbf{1.)} Image features are extracted from 10m/pixel Sentinel 2A data; \textbf{2.)} A subset of survey questions are selected considering their information overlap with the satellite image features; \textbf{3.)} Image features are combined with the subset of most important survey questions to predict the poverty rate.}
\label{fig:architecture}
\end{figure*}

\paragraph{\textbf{Contribution}}
In this work we seek to combine the benefits of both PMT based and satellite imagery based approaches and thus ask a central question: ``\textit{What additional information can be obtained from earth observation data to better complement PMTs when predicting poverty rates?}''. To this effect, we develop an approach to integrate image features and survey data to better estimate the poverty rate. Our contributions are three-fold:
\begin{itemize}
    \item This work shows that satellite image features are useful supplements to proxy means tests in estimating poverty rates. With ten survey questions, we show that including image features reduces the mean error in poverty rate estimates from 4.09\% to 3.88\%. At a larger number of survey questions (20 - 50), combining image features with survey questions always outperforms models that use only survey questions to estimate the poverty rate.
    \item We propose an approach for choosing survey questions that better complement image features. This approach achieves the best performance, reducing the poverty rate estimation error from 4.09\% to 3.71\%.
    \item We perform interpretability analysis to understand the features extracted from satellite imagery. We show that image features capture geographic differences in urbanization levels, which better complement survey questions that do \textit{not} highlight geographic differences. By exploiting image features to select a small set of survey questions, our approach maximizes the combined information captured in the small set of survey questions plus image features.
\end{itemize}
\section{Problem Formulation}
\label{problem}
For a geographic area, $a$, the poverty rate, $pr_a$, is defined as the proportion of households living below some poverty line\footnote{Poverty lines are typically set at a national level.}, \textit{pl}. Given the consumption expenditures of a household, $hce_i$, obtained from a survey, a binary household poverty indicator, $y_i$, is assigned using the criteria below:

\begin{equation*} 
y_i = \begin{cases} 
      \text{poor} & hce_i \leq pl \\
      \text{non-poor} & hce_i > pl 
      \end{cases}
\end{equation*}

Using this, $pr_a$ is set as the weighted average of $y_i$ for all households in the region:
\begin{equation}
    pr_a =\frac{\sum\limits_{i \in a} w_i y_i }{\sum\limits_{i \in a} w_i}
\end{equation}

Here the weight given to each household, $w_i$, are given by the survey used to calculate the household consumption expenditures.

We would like to develop a model, $\mathcal{F}$, that estimates a household poverty indicator, $\hat{y}_i$, as a function of a household's survey responses, $x_i$, and the satellite imagery features, $m_c$, collected from the same \textit{cluster}\footnote{A cluster is an anonymized location assigned to a group of households.}, $c$, that the household belongs to:
\begin{equation}
    \hat{y}_i = \mathcal{F}(x_i, m_c)
\end{equation}
The predicted poverty rate, $\hat{pr}_a$, can then be calculated as a weighted average of $\hat{y}_i$ over all households in a region.

Further, we would like to identify a subset of questions in a household survey that can be combined with cluster level satellite imagery features to maximize the performance of $\mathcal{F}$.

\section{Data} \label{data}
We use two datasets in this study, the Ethiopia Socioeconomic Survey (ESS), that consists of household level surveys used to determine poverty status, and Sentinel 2 imagery, that is used to derive household cluster level features.

\paragraph{Poverty rate}
The Ethiopia Socioeconomic Survey (ESS) dataset is a nationally representative household Living Standards Measurement Study (LSMS) carried out by the World Bank between 2015 to 2016 and is used to calculate the true poverty rate over different regions. This survey captures responses from 4954 households distributed over 528 clusters in Ethiopia~\citep{ess2015}.

Specifically, LSMS computed household consumption expenditures obtained from the survey responses are used to obtain the binary indicator of poor / non-poor households. Household consumption expenditure represents the value of goods and services purchased by a household during a given period. Consumption expenditures (adjusted to the household size) are compared to the national poverty line threshold of $\sim$ 3 USD/day (in 2011 dollars) where households with expenditures greater than the threshold are classified as non-poor.

\begin{figure}[b]
\centering
\includegraphics[width=0.5\textwidth]{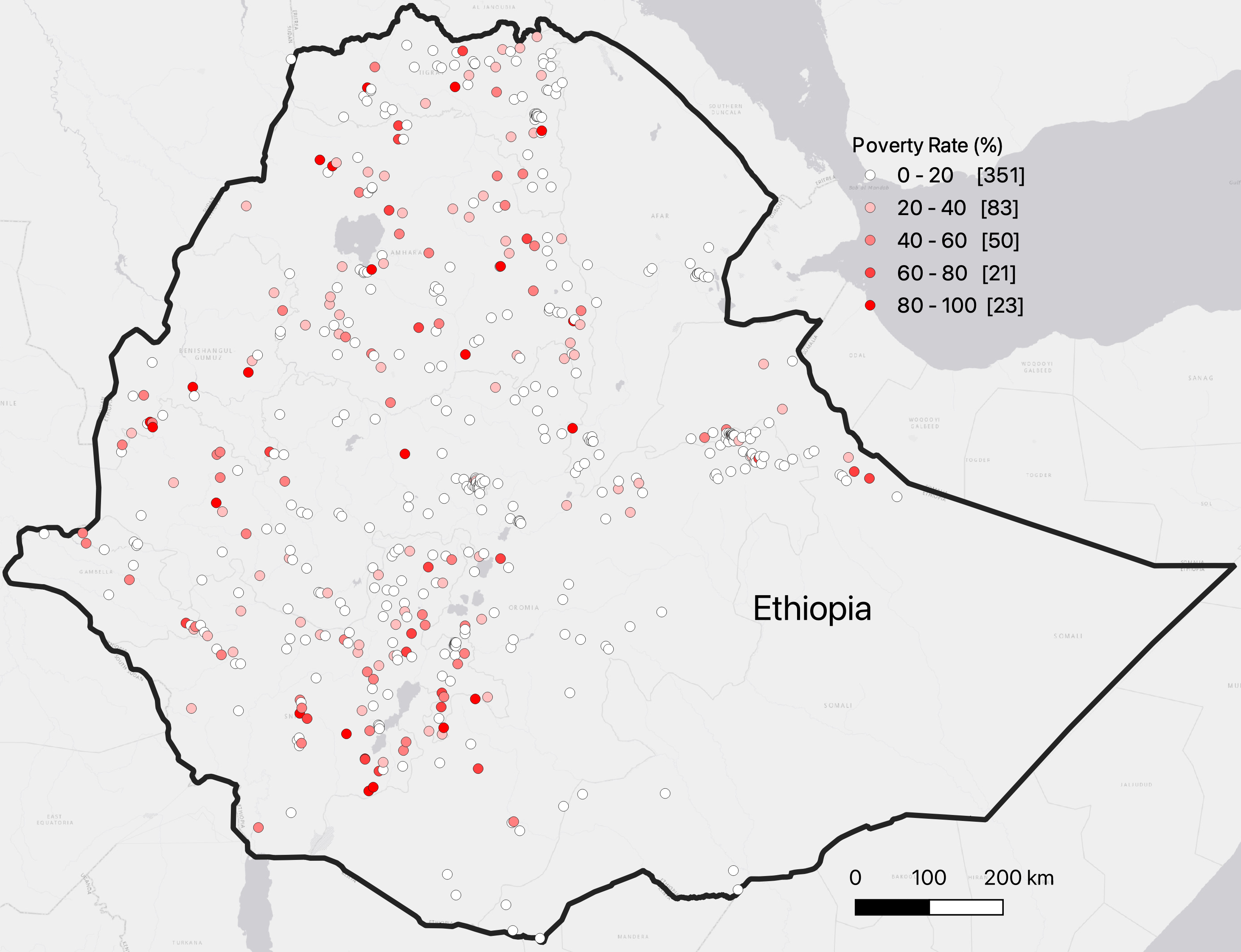}
\caption{Spatial distribution of the poverty rate over Ethiopia calculated from the 2015/2016 Living Standards Measurement Study.}
\label{fig:spatial_poverty_rate}
\end{figure}
Figure \ref{fig:spatial_poverty_rate} shows the spatial distribution of the poverty rate for geographic clusters within the dataset. Here, 66\% of the geographic clusters within the dataset have a poverty rate less than 20\% while 4.4\% have a poverty rate of over 80\%.

\paragraph{Survey questions}
The LSMS household survey responses from the above survey are used to develop the PMT. Categorical survey responses from the survey are converted to binary variables. Seasonal survey variables (e.g. consumption of seasonal foods) are excluded from the analysis together with variables with incomplete responses. A comprehensive list of survey questions can be accessed \textcolor{blue}{\href{https://microdata.worldbank.org/index.php/catalog/2783/data-dictionary}{here}}.

\paragraph{Sentinel-2 Imagery}
Sentinel-2 Level-2A data, is an open source 10-60 m/px spatial resolution satellite imagery product provided by the European Space Agency~\citep{sentinel2}. Sentinel-2 provides surface reflectance products in cloud optimized GeoTIFF format, accessible through Microsoft's Planetary Computer~\citep{microsoft_open_source_2022_7261897}. The LSMS survey period is used as the date query range to find cloud-free images for each of the geolocated clusters. For each location we select the lowest cloud cover image tile and download a 10 x 10 km crop of RGB imagery centered on the geolocated cluster.

\section{Multimodal Poverty Rate Prediction} 
\label{methods}
Our proposed method involves extracting relevant features from imagery and combining the image features with survey variables to estimate the poverty rate. In addition, we use the image features and survey variables to determine the right set of small survey questions to utilize in a proxy means test for estimating the poverty rate. Here we provide in depth descriptions of the methods used.

\paragraph{Image feature extraction.}
Image feature extraction is performed using the MOSAIKS featurization method proposed by ~\cite{rolf2021generalizable}. This featurization method is attractive as it does not require complex vision-based deep learning models, large compute resources, or hyper-parameter tuning to extract relevant image features. The empirical MOSAIKS approach is used, where a random set of 20 images are first selected from the training set. Of the randomly selected images, a random set of \textit{k}, 3 X 3 image subpatches are selected and these become the filters for feature extraction. Zero Component Analysis (ZCA) whitening are applied to the \textit{k} filters  as a image processing normalization step. The processed filters are then convolved with all images in the dataset and a RELU non-linear activation is applied.  Following this, an average pooling function is applied to the outputs to produce a scalar value for each filter \textit{k}. The resulting vector is the image feature vector which is used in subsequent analysis. This featurization step can be liken to a two layer non-trainable convolutional neural network with 27 parameters per filter.
In this work, 128 filters are extracted from images and used for subsequent analysis. By randomly selecting a small set of training images to build filters, this method provides a metric of similarity between images in the dataset and those selected as filters. This image feature extraction step is chosen over training an image based model because of the data-hungry nature of image models compared to the small number of images (528) in our dataset.

\paragraph{Selecting survey questions for a PMT}
A small set of relevant survey questions are identified using a two-step process: First, an Explainable Boosting Machine (EBM)~\cite{lou2012intelligible} is trained to predict the poverty indicator ($y$) using all survey variables. EBM is a tree-based gradient boosting algorithm that learns from only one feature at a time, building smaller models (known as feature functions) for each feature. To obtain a poverty prediction given a sample's input features, outputs from all feature functions are added, thus highlighting the contribution of each feature. EBM models combine the interpretability of Generalized Additive Models (GAMs)~\cite{hastie1990generalized} with the performance of gradient boosting models. 
After training the EBM model, the contribution of each survey variable to the prediction is measured and used to rank the survey variables. Given the survey feature importance scores, the \textit{n} most important survey variables are selected. A final step of retrieving other variables belonging to the same questions as the most important variables is performed. This process yields a reduced set of survey questions for the PMT and is called the \textit{survey guided} approach. 

The reduced set of questions are evaluated in a downstream task. Here, the small set of survey questions are used to train a model to predict the poverty indicator, $y_i$, of a household. The model performance is measured with a metric called poverty rate error -- the difference between the true poverty rate,\textit{pr}, and the predicted poverty rate,$\hat{pr}$.

To obtain survey questions that better complement image features when used in a PMT, we include image features in the survey question selection task. Survey variables and image features are first used to train a model to predict the poverty indicator. Feature importance scores are used to select the small set of survey questions to be used in the PMT. This approach is called the \textit{survey + image guided} approach. We compare the performance of the survey question selection methods when only survey variables are used (\textit{survey guided}) compared to when survey variables are combined with image features (\textit{survey + image guided}).

\paragraph{Data split}
The dataset of 4954 households is split into train and test sets to ensure out-of-sample performance is measured. Sixty-eight percent of the households are randomly placed in the train set while thirty-two percent are placed in the test set. Household allocation to train and test sets is done such that the poverty rate in each set is equal to the national poverty rate. This data split strategy is chosen to evaluate our model when a new set of random households are surveyed. The train dataset is used to determine the most salient survey questions. The test set is not used for variable selection but only used after the downstream task to evaluate the performance of the reduced set of survey variables in estimating the poverty rate.

\paragraph{Training and optimization.}
Both the survey variable selection model and the downstream poverty prediction model are trained with the households from the train set using k-fold cross-validation with $k=10$. The models are trained with a \textit{negative log-likelihood} loss. The LSMS survey sample weights are used during training to re-weight the model such that the model learns from underrepresented samples. A grid search hyper-parameter optimization over the learning rate (0.005, 0.0075, 0.01), number of leaves (10, 31, 50) and interactions (5, 10, 20) is performed using households in the training set during the survey variable selection step. The obtained hyper-parameters are used to evaluate the choice of variables in the downstream model. The results are reported on households from the out-of-sample test set. Uncertainty is measured by bootstrapping 100 samples from the test set 1000 times and computing the mean and standard deviations.

\paragraph{Metrics}
The main metric for evaluation is the poverty rate error (\textit{pre}). This error is the absolute difference between the true poverty rate and the predicted poverty rate.
\[\texttt{pre} = |pr - \hat{pr}|\]

The mean and standard deviations of \textit{pre} across the number of bootstrapping iterations (1000) is obtained and reported.

\section{Results and discussion} 
\label{results}
\paragraph{Evaluating the impact of image feature inclusion}
A key contribution of this work is the inclusion of image features with household survey variables to enchance proxy means testing. In this section, we evaluate the impact of image feature inclusion by measuring the poverty rate error when survey questions with and without image features are used in a PMT to estimate the poverty rate. 
\begin{figure}[b]
\centering
\includegraphics[width=0.45\textwidth]{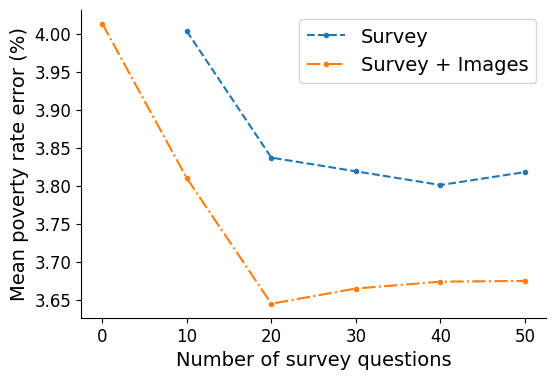}
\caption{Mean poverty rate (\%) error as a function of number of survey questions. Errors are lower when image features are included compared to when only survey questions are used.}
\label{fig:pmt_num_vars}
\end{figure}

\begin{table*}[ht]
\caption{Mean and standard deviation of the poverty rate error (PRE). Results are reported by \textit{variable selection method} and by PMT model input. The \textit{variable selection method} shows the set of features used to determine the top 10 survey questions input into the PMT. \textit{PMT model inputs} shows whether or not image features are combined with the top 10 questions to train the PMT model. Means of the PRE across 1000 iterations are reported with standard deviations shown in brackets.}
\begin{tabular}{ccccc}
\hline
\multirow{2}{*}{Variable selection method} & \multirow{2}{*}{PMT model inputs} & \multicolumn{3}{c}{Mean PRE (Std PRE)}  \\ \cline{3-5} 
                                                  &                             & All         & Rural       & Urban       \\ \hline
\multirow{2}{*}{Standard}                         & Survey                      & 4.09 (3.03) & 0.41 (0.30) & 1.07 ($\sim$ 0) \\
                                                  & Survey + Image              & 3.88 (2.94) & 0.32 (0.23) & 0.66 ($\sim$ 0) \\ \hline
\multirow{2}{*}{Survey Guided}                    & Survey                      & 4.00 (3.10) & 0.65 (0.34) & 0.36 ($\sim$ 0) \\
                                                  & Survey + Image              & 3.81 (2.92) & 0.63 (0.31) & 0.86 ($\sim$ 0) \\ \hline
\multirow{2}{*}{Survey + Image Guided}            & Survey                      & 3.95 (2.93) & 1.04 (0.36) & 0.70 ($\sim$ 0) \\
                                                  & Survey + Image              & 3.71 (2.84) & 0.30 (0.23) & 1.11 ($\sim$ 0) \\ \hline
\end{tabular}
\label{tab:pmt_pre_ppi_sg_sig}
\end{table*}

\begin{figure*}[!b]
\centering
\includegraphics[width=0.45\textwidth]{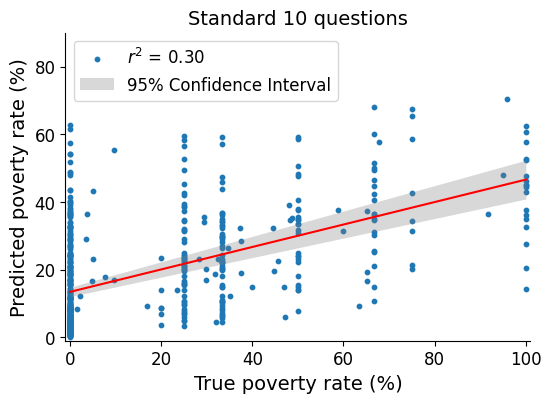}
\includegraphics[width=0.45\textwidth]{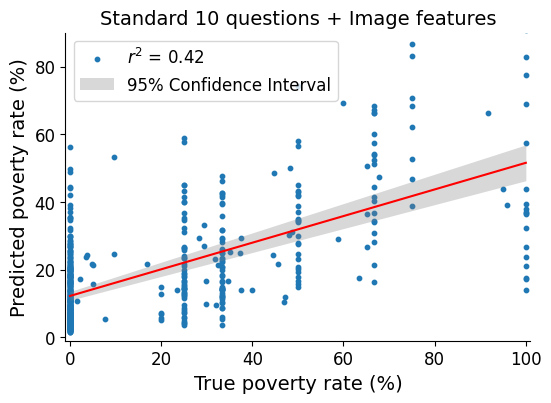}
\caption{Agreement between the true poverty rate and the predicted poverty rate at the cluster level when standard PMT questions are used compared to standard PMT questions + image features.}
\label{fig:clusterperformance}
\end{figure*}
Figure \ref{fig:pmt_num_vars} shows the mean poverty rate error over an increasing number of survey questions with and without the inclusion of image features. At zero survey questions (where only image features are used), the poverty rate error is highest though not much higher than when only ten survey questions are used. With the addition of a small number of survey questions (10) while retaining images, the poverty rate error decreases from 4.1 \% to 3.81 \%. Increasing the number of survey question to 20, reduced the poverty error rate with and without images, though larger gains are observed when images are included. Beyond 20 questions, adding more questions increases the poverty rate error when images are included in the PMT. However, the increased error observed with the inclusion of image features remains lower than the errors observed when only survey questions are used in the PMT.
In general, image features consistently improve the poverty rate estimation over using only survey variables. As more survey questions ($>$20) are acquired,  a slight decreases in the effectiveness of image features in estimating the poverty rate is observed.
\begin{figure*}[ht]
\centering
\includegraphics[width=0.75\textwidth]{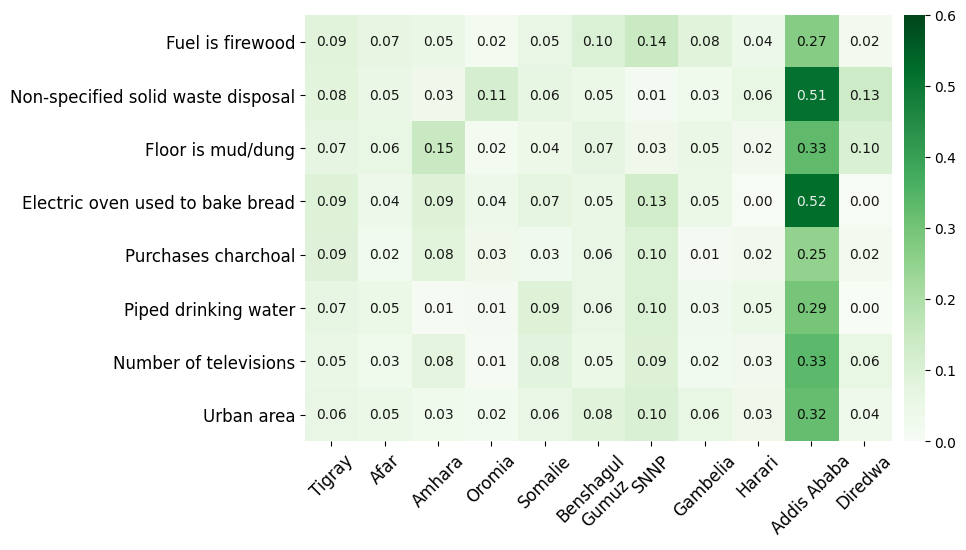}
\caption{Pearson correlation between the regional survey indicators and survey indicators that highly correlate with image features. Survey features that highly correlate with image features also correlate with households being in the urban region of Addis Ababa. This suggests that image features capture the geographic differences between urban and non-urban areas.}
\label{fig:correlation_image_geosurvey_feats}
\end{figure*}
\paragraph{Evaluating the performance of the survey variable selection task}
The survey variable selection process is evaluated by measuring the poverty rate error of a PMT when the reduced set of survey questions are used. Here, the error in the poverty rate obtained with the set of 10 survey questions from our variable selection approach is compared to the error in the poverty rate when a standard set of PMT questions are used. The Poverty Probability Index (PPI) is a poverty measurement tool that uses 10 questions about household characteristics and asset ownership in a PMT to compute the likelihood of being poor\cite{ppi}. In Ethiopia, 10 standard questions are used to compute the PPI.

Table \ref{tab:pmt_pre_ppi_sg_sig} shows the mean poverty rate error and in brackets the standard deviations when the 10 standard questions are used. When images are combined with the 10 standard questions for Ethiopia, improvements in poverty rate estimation are observed as the error means and standard deviations in the poverty rate decrease from 4.09 \% to 3.88 \% and 3.03 \% and 2.94\%, respectively. This observation is consistent across the full test set and by urbanization level.\footnote{Note that the standard deviations in the poverty error rate are very small for urban regions and as a result are reported as approximately zero.}

The variable selection method is performed in two ways: using only survey variables (\textit{survey guided}) and using survey variables with image features (\textit{survey + image guided}). 

The selected questions obtained through the \textit{survey guided }variable selection mechanism provides improved estimates of the poverty rate across the whole dataset when only survey questions are used in the PMT. The gains are obtained with improved performance in the urban set. Inclusion of images features to the survey questions obtained through the \textit{survey guided} mechanism further reduces the overall poverty rate errors from 4.09 \% to 3.81\%, compared to 3.88 \% when the standard questions are used. Inclusion of image features to the survey questions obtained from the \textit{survey guided} variable selection process compared to inclusion of image features to the standard questions shows that images are helpful in reducing the errors in the overall poverty rate estimate but may not be tuned to performing better within a specific urbanization region. This suggests that image features capture the differences between urbanization levels and geographies but do not extract sufficient differences within the same urbanization region. This behavior may be a limitation of the images used as they are at a medium resolution of 10 m/pixel, and thus may not capture adequate differences within geographies (e.g. differences in rooftop materials within the same region or image scene) that are more evident at higher resolutions. 

The \textit{survey + image guided} variable selection mechanism performs the best in estimating the overall poverty rate when image features are combined with the small set of obtained survey variables. The overall poverty rate errors decrease from 4.09 \% to 3.71 \% while the rural poverty rate errors decrease from 0.32 \% to 0.30 \%. Under the \textit{survey + image guided}, the poverty rate errors increase from 0.66 \%  to 1.11 \%. The improved performance for the overall and rural set suggests that by guiding the variable selection process with both image features and survey variables, complementary survey variables are selected to maximize information for improved performance. In addition to decreased means in the poverty rate error, the \textit{survey + image guided} mechanism also shows the lowest standard deviations in the poverty rate errors. 

\paragraph{Cluster level performance}
Using the ten standard questions, poverty rate estimates are also evaluated at the cluster level for the test set. Figure \ref{fig:clusterperformance} shows the agreement between the true poverty rates and the predicted poverty rates with and without image features. Here each point represents a geographic cluster. $r^2$ values are reported at the top left corner of each sub-figure. The line of best fit is also shown along side the 95 \% confidence interval of a linear regression model. Adding image features to the standard questions increases the $r^2$ to 0.42 from 0.30 when image features are not used. When \textit{survey + image guided} variables are used, the $r^2$ increases to 0.45. These findings suggest that inclusion of image features from the single-step featurization method supports better estimates of the poverty rate at varying spatial resolutions from higher resolution geographic clusters to lower resolution nationwide estimates.

\paragraph{Inspection of image features}
Image feature inspection is important to understanding the visual characteristics that are retrieved during image feature extraction and how these characteristics relate to survey questions. The non-trainable nature of the featurization approach does not support direct extraction of spatial characteristics that may correlate with poverty rates, as result we explore an alternative approach described in this section. Principal component analysis (PCA) is first applied to the image features to reduced the image feature vector from 128 to its top three most relevant components. This dimensionality reduction step supports the interpretability of image features while preserving the relevant extracted information. Pearson correlations between the principal components and the survey variables are computed as an approximation of the information captured by images. Image features that highly correlate with survey variables provide some interpretation of the image features. The following ten survey questions had the highest Pearson correlation with the top three components of the reduced image features: urban area (urban/rural), whether the household is in Addis Ababa, whether the household is in Gambelia, the use of Mud/Dung as the predominant floor material, the cooking fuel, the use of piped water into the yard as the main source of drinking water, whether the household purchases charcoal, the use of an electric oven, the number of televisions owned by the house and the used of non-specified solid waste disposal. Of the set of most correlated survey variables, survey variables such as regional and urbanization differences are more observable from imagery than others. To further understand these correlations, these ten survey variables are correlated with geographic survey variables to better understand their spatial characteristics.

Figure \ref{fig:correlation_image_geosurvey_feats} shows the Pearson correlations between the ten survey variables that agree with images (shown on the y-axis) and the regional survey variables (x-axis). Addis Ababa and Gambelia are omitted from the y-axis as correlations to themselves are 1. Amongst the remaining eight survey variables with the highest image feature correlations, high correlations with the Addis Ababa regional indicator are also observed.

This observation highlights two findings. First, while image features may be correlated with less observable variables such as water source and cooking material, these correlations emerge because image features capture the underlying geographic differences that the less observable variables encode.
Secondly, image features from the single-step featurization strongly encode the differences between urban (Addis Ababa) and non-urban regions. This second observation further explains why the inclusion of image features improves overall estimates of the poverty rate across rural and urban regions but does not perform as well within an urbanization region. 

\paragraph{Complementary question discovery}
The ten standard and \textit{survey guided} questions consist of a regional indicator which encodes the regions within the survey. In the \textit{survey + image guided} approach, the regional indicator is excluded (as the image features capture regional and geographic differences across clusters) and the survey variable selection model is allowed to chose the ten most relevant non-geographic survey questions. Through this process the \textit{survey + image guided} approach retains five out of the ten questions initially chosen by the \textit{survey guided }approach.
Five new complementary survey questions are discovered when image features and survey question are used to determine the most important survey questions, thereby maximizing the available information to estimate a poverty rate.

\section{Conclusion}
In this paper, we propose an approach to predicting the poverty rate of a region through the combination of survey variables and satellite image features. We show that the inclusion of image features consistently improves the estimation of the poverty rate over using survey questions alone. 

We also proposed a method for selecting the most relevant survey questions for a PMT when image features are available. Our \textit{survey + image guided} approach outperforms other survey question section methods for poverty rate estimation. Additionally, we conducted an interpretability analysis around the visual features extracted by the single-step featurization method. We found that the satellite image features capture underlying geographic differences in urbanization levels complementing the survey questions for improved overall estimation of the poverty rate. 

Overall we demonstrate that satellite image features provide additional information for poverty rate prediction. This better supports the possibility of using a cost effective smaller set of survey variables as is the case with proxy-means tests while keeping reasonable poverty predictions. 
\bibliographystyle{ACM-Reference-Format}
\bibliography{compass23-3}
\appendix

\end{document}